%% file: main.tex
\documentclass[10pt,twocolumn,letterpaper]{article}

\usepackage[pagenumbers]{cvpr} % To force page numbers, e.g. for an arXiv version

\usepackage{cvpr}      % To produce the REVIEW version
\usepackage{multirow}
\usepackage{colortbl}
\usepackage{booktabs}
\usepackage{makecell}
\usepackage{adjustbox}
\usepackage{booktabs}
\usepackage{graphicx}
\usepackage[table]{xcolor}
\usepackage[table]{xcolor}

\definecolor{rankfirst}{RGB}{169,52,50}
\definecolor{ranksecond}{RGB}{40,97,138}
\definecolor{rankthird}{RGB}{71,122,96}

\newcommand{\best}[1]{\textcolor{rankfirst}{\textbf{#1}}}
\newcommand{\second}[1]{\textcolor{ranksecond}{\underline{#1}}}
\newcommand{\third}[1]{\textcolor{rankthird}{\textit{#1}}}
\input{preamble}
\definecolor{cvprblue}{rgb}{0.21,0.49,0.74}
\usepackage[pagebackref,breaklinks,colorlinks,allcolors=cvprblue]{hyperref}

\def\confName{CVPR}
\def\confYear{2027}
\title{PrismRestore: A Compact Restoration Toolbox for Agentic Image Restoration}
\author{
Zhaocheng Yu \quad
Kui Jiang \quad
Jing Cao \quad
Yiang Chen\\
Xiaocheng Feng \quad
Junjun Jiang \quad
Xianming Liu\\
Harbin Institute of Technology\\
{\tt\small
\{yuzhaocheng,caojing,chenya\}@stu.hit.edu.cn}\\
{\tt\small
\{jiangkui,jiangjunjun,csxm\}@hit.edu.cn,
xcfeng@ir.hit.edu.cn}
}
\begin{document}
\maketitle
\input{sec/0_abstract}    
\input{sec/1_intro}

\input{sec/2_related_work}
\input{sec/3_method}
\input{sec/4_experiment}

\input{sec/5_conclusion}
\input{sec/LLM_Usage_Statement}
{
    \small
    \bibliographystyle{ieeenat_fullname}
    \bibliography{main}
}


\end{document}

%% file: sec/0_abstract.tex
\begin{abstract}
% Agentic image restoration adapts to mixed degradations by composing specialized tools, but typically relies on a collection of separately parameterized restoration models. 
% Can a single frozen all-in-one model instead provide a controllable toolbox? We observe that different restoration tasks induce distinct yet overlapping channel responses within a shared model. Motivated by this observation, we propose a compact agentic framework (PrismRestore) that constructs multiple callable restoration actions through task-specific channel modulation.
% An offline task-prior search groups channels using activation and gradient statistics, preserves representative responses, and searches modulation coefficients for non-representative channels.
% The resulting actions share a frozen backbone and differ only in their modulation coefficients.
% We then train a state-conditioned policy with Group Relative Policy Optimization to select actions as the image evolves.
% A quality-grouped, process-aware objective normalizes fidelity and perceptual feedback separately within each trajectory group before advantage fusion, and complements terminal evaluation with return-to-go computed from stepwise quality changes.
% Experiments on synthetic composite-degradation and real-world multi-task benchmarks show improved restoration quality over the evaluated all-in-one baselines and competitive quality against multi-model agents, with lower restoration-toolbox storage requirements and higher inference efficiency than the evaluated agentic baselines.

Agentic image restoration composes operations to handle mixed degradations, yet its toolbox often requires storing multiple restoration networks.
We propose PrismRestore, which constructs a compact bank of callable restoration actions from a single frozen all-in-one model.
Our approach builds on task-dependent yet overlapping channel responses in the shared model.
An offline task-prior search uses activation and gradient statistics to group channels with similar response profiles, then searches task-specific modulation coefficients over non-representative channels while leaving representative channels unmodulated.
The resulting prior bank enables explicit action selection and composition without additional restoration backbones or task-specific adapter modules.
A state-conditioned policy learns to compose these actions using Group Relative Policy Optimization.
Its objective separately normalizes fidelity and perceptual feedback within groups of trajectories sampled for the same input, and combines terminal evaluation with return-to-go derived from stepwise quality changes.
Experiments on synthetic composite-degradation and real-world multi-task benchmarks show improvements over the evaluated all-in-one baselines and restoration quality competitive with multi-model agents.
PrismRestore achieves these results with a single restoration backbone and faster inference than the evaluated agentic baselines.
\end{abstract}

%% file: sec/1_intro.tex
\section{Introduction}
\label{sec:intro}

% For images affected by mixed degradations, the effectiveness of a restoration operation depends on the image state in which it is applied. Noise, blur, haze, and low-light artifacts can coexist, and each restoration operation changes the input encountered by subsequent operations~\cite{chen2024restoreagent,zhu2025agenticir}. An effective restoration sequence should therefore account for the evolving image state, rather than rely solely on degradation labels assigned to the initial observation. This motivates adaptive composition of restoration operations for complex image restoration.

% Images can be affected by multiple degradations, such as noise, blur, haze, and low-light artifacts. In sequential restoration, each operation changes the image presented to subsequent operations~\cite{chen2024restoreagent,zhu2025agenticir}, so its effectiveness depends on the current image state. An effective restoration policy should therefore select and compose operations according to the evolving image, accounting for the effects of preceding restoration steps.
Images captured in uncontrolled environments often contain mixtures of noise, blur, haze, and low-light artifacts.
Sequential restoration addresses such mixtures by composing operations, each of which changes the input encountered by subsequent steps~\cite{chen2024restoreagent,zhu2025agenticir}.
Consequently, the usefulness of a restoration operation depends on the current image state and its position in the sequence.
This motivates policies that adapt their actions as restoration proceeds.
Realizing such policies requires a toolbox whose actions can be explicitly selected and composed at a manageable storage cost.
% Deep image restoration has progressed from degradation-specific models~\cite{zhang2017dncnn,zamir2020mirnet,zamir2021mprnet,zamir2022restormer,chen2022nafnet} to all-in-one networks that handle multiple restoration tasks within a unified architecture.
% However, early all-in-one methods were largely trained on single-degradation data, while real-world images often suffer from multiple interacting degradations.
% Recent approaches alleviate this limitation by training on predefined composite degradations; however, their fixed feed-forward inference provides limited flexibility in explicitly adapting restoration operations and their execution order to evolving image states, particularly when encountering complex or unseen degradation combinations.
% Agentic restoration therefore formulates complex restoration as a sequential decision process, where an agent analyzes the evolving image state and invokes task-specific tools accordingly.
% Nevertheless, existing systems typically rely on numerous independently trained models, introducing considerable deployment overhead and fragmented knowledge across heterogeneous tools.

\begin{figure}[t]
  \centering
   \includegraphics[width=0.98\columnwidth]{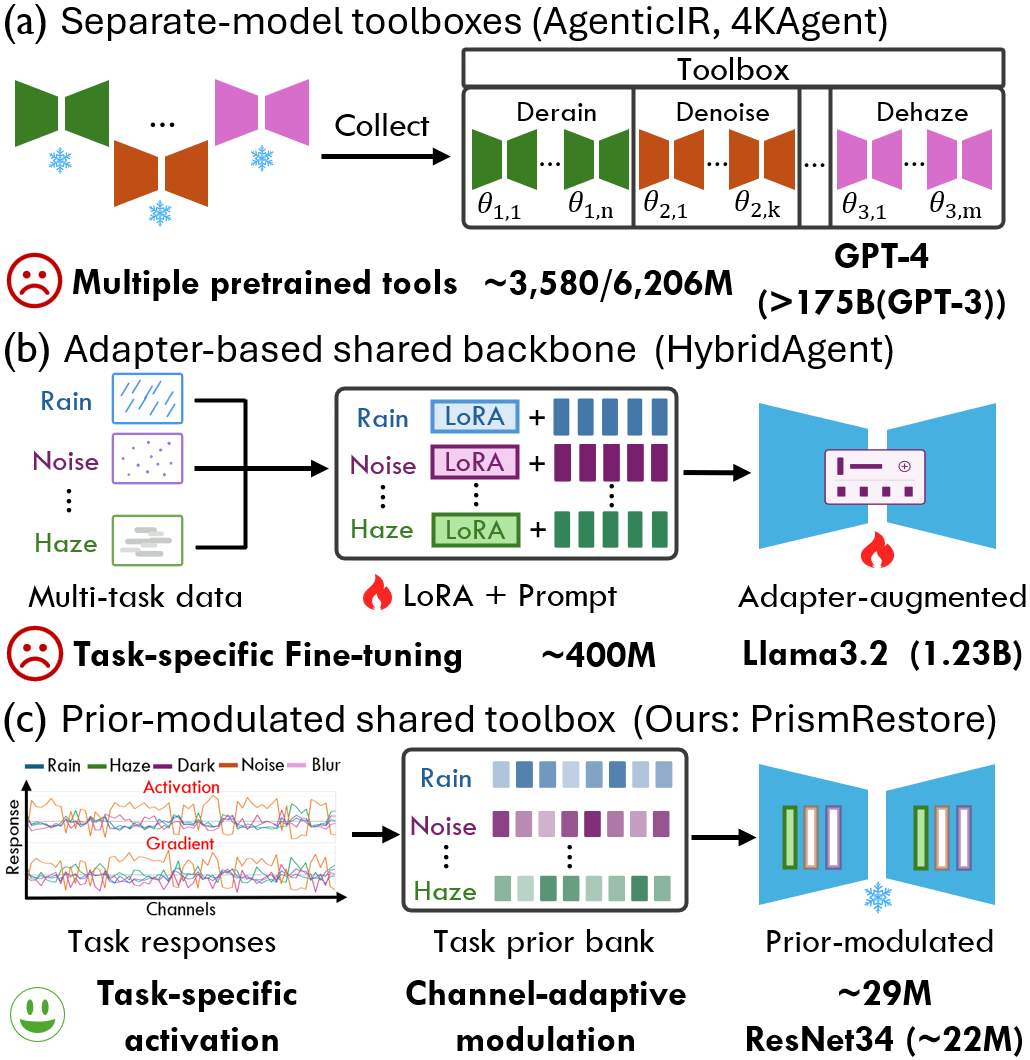}
   \caption{Parameterization of restoration toolboxes.
(a) Independent tools store separate restoration networks.
(b) Adapter-based tools share a backbone with task-specific
parameters. (c) PrismRestore constructs restoration actions
using compact modulation priors on a frozen all-in-one
backbone. The two parameter counts indicate toolbox
and planner parameters, respectively.
   % (a) Separate models incur substantial storage costs. (b) Backbone sharing still requires task-specific adapter parameters. (c) PrismRestore activates restoration modes within a frozen backbone using compact task priors, substantially reducing toolbox storage.
   }
 \vspace{-4mm}
   \label{fig:onecol}
\end{figure}

Deep image restoration has progressed from task-specific models~\cite{zhang2017dncnn,zamir2020mirnet,zamir2021mprnet,zamir2022restormer,chen2022nafnet} to all-in-one networks that handle multiple degradations within a shared architecture~\cite{potlapalli2023promptir,conde2024instructir}. 
Agentic systems further enable explicit control over restoration sequences through degradation assessment, planning, and tool invocation~\cite{chen2024restoreagent,zhu2025agenticir}.
Recent work advances this control through policy optimization~\cite{lu2025restorer1,jia2026tiragent} and joint optimization of planning and execution~\cite{zhu2026opera}.
Alongside these advances in decision-making, the way restoration actions are parameterized determines the cost of maintaining the toolbox.

Many restoration agents instantiate their actions with separately parameterized models~\cite{chen2024restoreagent,zhu2025agenticir,lu2025restorer1}, requiring a collection of full restoration networks (Fig.~\ref{fig:onecol}(a)).
HybridAgent~\cite{li2026hybrid} reduces backbone duplication through task-specific LoRA and prompt tuning, while retaining learned adaptation parameters for individual tools (Fig.~\ref{fig:onecol}(b)).
This motivates a more compact action representation:
\textit{can an existing all-in-one model serve as a restoration toolbox through task-specific channel modulation, with its pretrained weights kept frozen and no additional adapters?}
Our analysis of activation and gradient statistics reveals distinct yet overlapping channel response profiles across restoration tasks within the same all-in-one model.
Task-dependent variation provides a basis for controlling restoration behavior, while the overlap suggests that some responses support multiple tasks.
Aggressive channel suppression may therefore discard information useful beyond the target task.
The challenge is to identify a compact modulation space that differentiates restoration actions while limiting disruption to shared computation.

% To address this challenge, we construct a shared-backbone toolbox through an \emph{offline task-prior search}. 
% Activation and gradient statistics characterize task-conditioned channel behavior. We group channels with similar response profiles, preserve representative responses, and search degradation-specific modulation coefficients over non-representative channels identified by response-based clustering. 
% The resulting task-specific coefficient sets form a prior bank, with each prior defining a callable restoration action on the same frozen backbone.
% Thus, the toolbox stores one restoration model and its task-specific modulation coefficients, rather than a separate restoration network for each action. The search is performed offline; at inference, the agent selects a task prior and applies its modulation coefficients to the shared frozen backbone,
% % invokes the resulting actions by selecting their corresponding coefficients, 
% as shown in Fig.~\ref{fig:onecol}(c).
% To address this challenge, we propose PrismRestore, an agentic image restoration framework that constructs a shared-backbone toolbox through  \emph{offline task-prior search}.
To this end, we propose \textbf{PrismRestore}, an agentic restoration framework that constructs a compact toolbox from a single frozen all-in-one model.
% To address this challenge, we construct a shared-backbone toolbox through \emph{offline task-prior search}.
An \emph{offline task-prior search} first characterizes channels using task-conditioned activation and gradient statistics and groups channels with similar response profiles.
It then searches degradation-specific modulation coefficients over non-representative channels, leaving representative channels unmodulated.
The resulting coefficient sets form a prior bank, where each prior defines a selectable restoration mode on the shared backbone.
At inference, a policy selects and applies these precomputed priors to invoke restoration actions (Fig.~\ref{fig:onecol}(c)).
The toolbox stores only the pretrained restoration backbone and its compact priors, without task-specific adapters.
% Activation and gradient statistics characterize task-conditioned channel behavior. We group channels with similar response profiles, preserve representative responses, and search degradation-specific modulation coefficients over non-representative channels identified by response-based clustering. 
% The resulting task-specific coefficient sets form a prior bank, with each prior defining a callable restoration action on the same frozen backbone. 
% The toolbox thus stores one restoration model and compact modulation priors, without separate networks or task-specific adapter modules. 
% At inference, the policy selects a prior to activate the corresponding restoration mode, as illustrated in Fig.~\ref{fig:onecol}(c).

To compose these actions, we train a policy conditioned on the current image state using Group Relative Policy Optimization (GRPO)~\cite{shao2024deepseekmath}.
Directly aggregating fidelity and perceptual feedback makes their relative influence sensitive to metric scale.
Following prior work on decoupled multi-metric advantages~\cite{liu2026gdpo,jia2026tiragent}, we normalize these feedback signals separately within groups of trajectories sampled for the same input before advantage fusion.
We also draw on stepwise quality feedback~\cite{lu2025restorer1} to complement terminal evaluation with return-to-go computed from quality changes after successive actions.

Our contributions are summarized as follows:
\begin{itemize}
    \item We introduce a compact restoration toolbox built from a single frozen all-in-one model. Response-guided task-prior search constructs multiple selectable restoration actions through channel modulation, requiring only one stored backbone and compact per-task coefficients.
    %We construct multiple callable restoration actions from a single frozen all-in-one model through offline task-prior search. Task-conditioned channel modulation provides an explicit action interface without maintaining independently parameterized restoration backbones.
%We propose a compact agentic image restoration framework based on a shared-parameter toolbox. Through offline task-prior search, the implicit degradation-specific capabilities of a frozen all-in-one model are transformed into explicit and controllable restoration actions, which are sequentially scheduled by a GRPO-trained policy according to the evolving image state. This design avoids assembling multiple independent restoration models and substantially reduces parameter redundancy and deployment overhead.

    \item We adapt group-relative policy learning to compose these actions according to the evolving image state, combining separately normalized fidelity and perceptual feedback with terminal evaluation and return-to-go from stepwise quality changes.
%We introduce a decoupled process-aware reward for multi-step restoration policy learning. Fidelity and perceptual quality are independently normalized within each trajectory group before advantage fusion, reducing interference between the two objectives. In addition, terminal trajectory evaluation is complemented with return-to-go accumulated from stepwise quality changes, providing more informative supervision for the contribution of individual actions.

    \item Experiments on synthetic composite-degradation and real-world multi-task benchmarks demonstrate improvements over the evaluated all-in-one baselines and restoration quality competitive with multi-model agents. PrismRestore uses a single restoration backbone and achieves faster inference than the evaluated agentic baselines.
\end{itemize}

%-------------------------------------------------------------------------

%% file: sec/2_related_work.tex
\section{Related Work}

%-------------------------------------------------------------------------
\subsection{All-in-One Image Restoration}

All-in-one image restoration aims to handle diverse degradation types within a unified model~\cite{chen2021ipt,valanarasu2022transweather,li2022airnet,park2023adms,wang2023promptrestorer,potlapalli2023promptir}.
Early approaches learned degradation representations to guide a shared restoration network, while subsequent studies introduced degradation-aware prompts to adapt restoration features to different corruptions.
DA-CLIP~\cite{luo2024daclip} incorporates vision-language priors through degradation and content embeddings, InstructIR~\cite{conde2024instructir} enables restoration control using natural-language instructions, and AutoDIR~\cite{jiang2024autodir} automatically identifies unknown degradations and restores images using a latent diffusion model.
AdaIR~\cite{cui2025adair} instead exploits degradation-dependent frequency characteristics to adaptively modulate restoration features. 
Moving beyond isolated degradations, OneRestore~\cite{guo2024onerestore} constructs composite-degradation data and performs controllable restoration with scene descriptors. 
Despite these advances, existing all-in-one methods generally lack explicit mechanisms for selecting and composing restoration operations according to evolving image states, limiting their flexibility in handling complex degradation combinations.

% Despite these advances, existing all-in-one methods largely rely on fixed feed-forward inference and predefined training distributions, limiting their ability to adaptively determine restoration operations and execution orders for complex or unseen degradation combinations.

%-------------------------------------------------------------------------
\subsection{Agent-Based Image Restoration}
% However, their tool libraries are commonly assembled from independently trained task-specific models, increasing parameter redundancy and deployment cost while introducing heterogeneous restoration knowledge across tools.
Agent-based image restoration formulates complex
restoration as a sequential decision-making problem,
enabling adaptive selection and composition of
restoration operations.
Early work such as RL-Restore~\cite{yu2018rlrestore}
employs reinforcement learning to construct
restoration toolchains.
Recent methods extend this paradigm through more
advanced planning and coordination.
RestoreAgent~\cite{chen2024restoreagent} and
AgenticIR~\cite{zhu2025agenticir} leverage multimodal
reasoning for degradation assessment and tool selection,
while MAIR~\cite{jiang2026mair} and 4KAgent~\cite{zuo20254kagent}
organize specialized restoration tools through
multi-agent collaboration and iterative planning.
HybridAgent~\cite{li2026hybrid} integrates fast, slow,
and feedback agents for efficient user interaction and
incorporates mixed-degradation restoration to reduce
error propagation, while SEAR~\cite{cui2026sear} combines
deliberate tree-search planning with memory-guided execution.

\begin{figure*}[!htbp]
  \centering
   \includegraphics[width=0.98\textwidth]{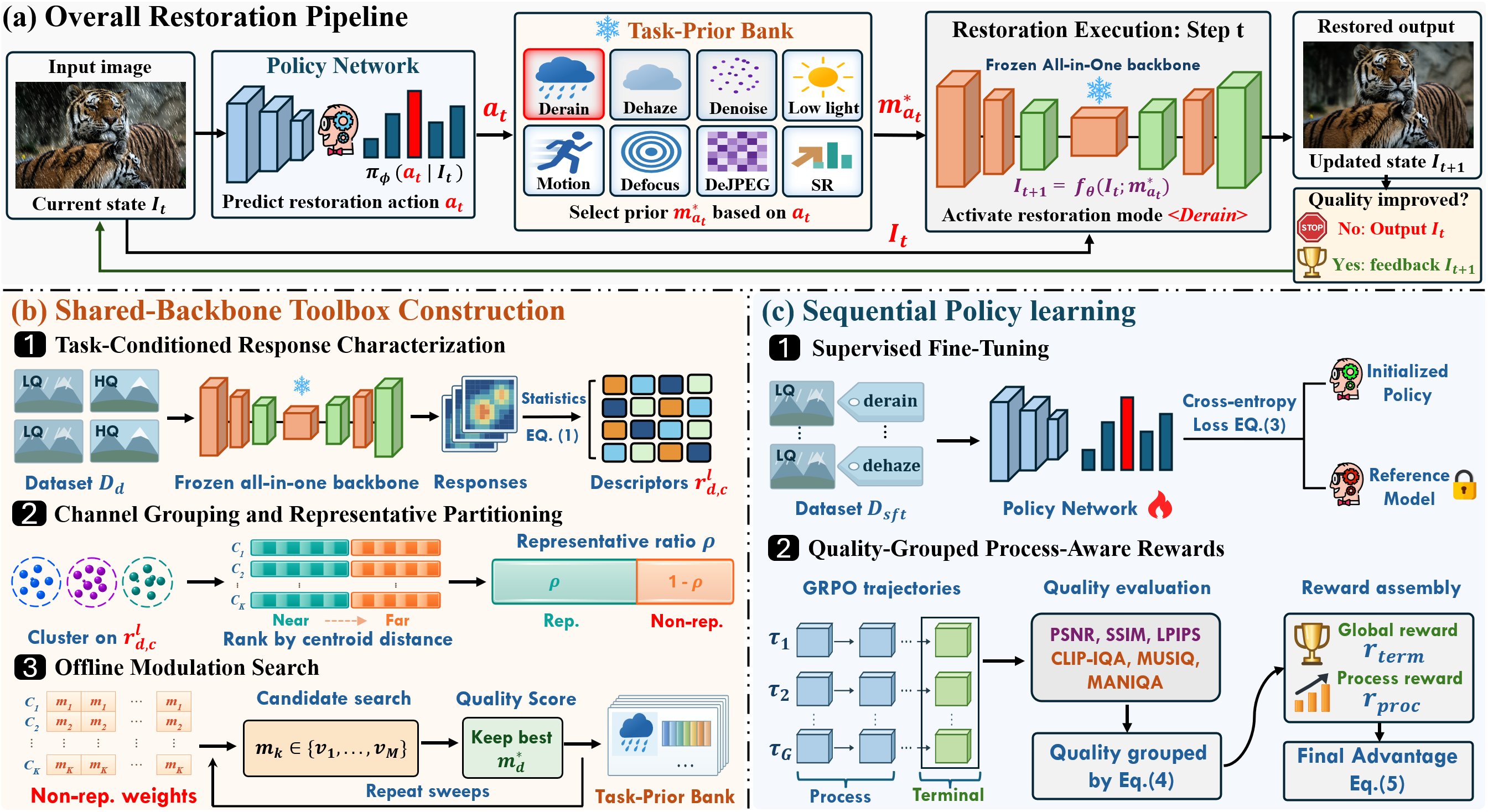}
   \caption{Overview of PrismRestore. (a) Overall restoration pipeline with state-conditioned action selection and quality-guided stopping. (b) Shared-backbone toolbox construction through channel response characterization, representative partitioning, and offline modulation search. (c) Sequential policy learning with SFT initialization and quality-grouped process-aware rewards.}
   \vspace{-4mm}
   \label{pipeline}
   % \vspace{-2mm}
\end{figure*}
More recent studies investigate learned restoration
policies and iterative refinement.
TIR-Agent~\cite{jia2026tiragent} employs supervised
initialization and reinforcement learning to improve
sequential tool selection, while
RAR~\cite{chen2026rar} integrates restoration
and quality assessment for iterative image enhancement.
FAPE-IR~\cite{liu2026fape} combines multimodal planning
with a frequency-aware diffusion executor.
Despite these advances, restoration planning and
toolbox construction are often treated as separate
components, with many approaches relying on
collections of specialized models or task-specific
adapters.
In contrast, we derive task-conditioned restoration
actions through offline channel modulation of a
single frozen all-in-one backbone and learn to
compose them using quality-grouped fidelity and perceptual
feedback with stepwise return-to-go.

%% file: sec/3_method.tex
\section{Method}
\label{sec:method}

\subsection{Restoration Pipeline}
As illustrated in Fig.~\ref{pipeline}(a),
our framework performs iterative restoration through
a state-conditioned policy and a shared-backbone toolbox.
Given the current image $I_t$, the policy network predicts
a distribution over the available restoration actions
and selects $a_t$.
The corresponding task-specific modulation prior
$m_{a_t}^{*}$ is retrieved from the preconstructed
task-prior bank and applied to the frozen all-in-one
backbone $f_\theta$, producing the updated image
$I_{t+1}$.
% $I_{t+1}=f_\theta(I_t;m_{a_t}^{*})$.
The updated image is then fed back to the policy
for subsequent action selection, allowing restoration
operations to be adaptively composed according to
the evolving image state.
During inference, we employ a quality-guided stopping
criterion based on the no-reference perceptual score
$Q$ defined in Sec.~\ref{sec:policy_learning}.
After each restoration action, we compare the perceptual
quality of the updated image with that of the preceding
state.
If $Q(I_{t+1}) \leq Q(I_t)$, the current restoration result
is discarded, and $I_t$ is returned as the final output.
Otherwise, restoration continues until the maximum
of $H=6$ steps is reached. Additional method details are provided in the supplementary material.
% This mechanism enables image-dependent restoration
% depths without introducing an additional stopping
% action into the policy.

\subsection{Shared-Backbone Toolbox Construction}
\label{sec:toolbox_construction}

As illustrated in Fig.~\ref{pipeline}(b), we construct a compact restoration toolbox by extracting
task-conditioned restoration behaviors from a frozen
all-in-one backbone. For each restoration task $d$,
an offline search is performed on the corresponding
paired single-degradation dataset $\mathcal{D}_d$.
The construction consists of three stages:
task-conditioned response characterization,
channel grouping and representative partitioning,
and offline modulation search. Throughout this process,
the backbone parameters $\theta$ remain fixed.

\noindent\textbf{Task-Conditioned Response Characterization.}
Different restoration tasks exhibit distinct yet
overlapping channel responses within the shared backbone.
To characterize these task-dependent behaviors, we collect
activation and gradient statistics from candidate modulation
layers. Let $Z^l\in\mathbb{R}^{C_l\times H_l\times W_l}$
denote the feature map at layer $l$. For channel $c$ of
sample $n$, we compute its activation magnitude
$p_{n,c}^l$ and gradient-weighted response $q_{n,c}^l$,
and aggregate them into a task-conditioned descriptor:
\begin{equation}
\begin{aligned}
p_{n,c}^{l}
&=\sqrt{\frac{1}{H_lW_l}
\sum_{h,w}(Z_{n,c,h,w}^{l})^2},\\
q_{n,c}^{l}
&=-\frac{1}{H_lW_l}\sum_{h,w}
Z_{n,c,h,w}^{l}
\frac{\partial\mathcal{L}_{rec}}
{\partial Z_{n,c,h,w}^{l}},\\
\mathbf r_{d,c}^{l}
&=[\mu(p_c^l),\sigma(p_c^l),
\mu(q_c^l),\sigma(q_c^l)].
\end{aligned}
\label{eq:channel_response}
\end{equation}
Here, $\mathcal{L}_{rec}$ is an $\ell_1$ reconstruction
loss used only for response characterization, while
the mean and standard deviation are computed over
samples from $\mathcal{D}_d$. The resulting descriptors
are standardized and $\ell_2$-normalized before
clustering, jointly encoding forward activation
characteristics and gradient-based responses.

\noindent\textbf{Channel Grouping and Representative Partitioning.}
For each task and candidate layer, we apply K-means
to group channels with similar response descriptors,
followed by quota-based balanced reassignment to avoid
highly imbalanced clusters. Within each cluster
$\mathcal C_{d,k}^{l}$, channels are ranked by their
distances to the cluster centroid. Given a representative
ratio $\rho$, the
$\lceil(\rho)|\mathcal C_{d,k}^{l}|\rceil$
channels closest to the centroid form the representative
subset $\mathcal C_{d,k}^{l,\mathrm{rep}}$, while the
remaining channels constitute the non-representative subset
$\mathcal C_{d,k}^{l,\mathrm{non}}$.
This partition preserves the characteristic
response patterns captured by representative
channels, while enabling task-specific modulation
of non-representative channels to limit
perturbations to these patterns.

% This partition retains representative channel responses
% while restricting subsequent task-specific modulation
% to the non-representative subset, reducing interference with
% shared representations.

\noindent\textbf{Offline Modulation Search.}
We search task-specific modulation coefficients over
the identified non-representative channels while keeping the
representative channels unchanged.
For task $d$, the non-representative channels within cluster
$k$ at layer $l$ share a modulation coefficient
$m_{d,k}^{l}$, yielding
\begin{equation}
\widetilde{Z}_{d,c}^{l} =
\begin{cases}
m_{d,k}^{l} Z_c^{l},
& c \in \mathcal{C}_{d,k}^{l,\mathrm{non}},\\
Z_c^{l},
& \mathrm{otherwise}.
\end{cases}
\label{eq:channel_modulation}
\end{equation}

For each task, candidate modulation coefficients are
evaluated on the corresponding single-degradation
search set using a composite quality criterion that
considers both reconstruction fidelity and perceptual
quality. The coefficient configuration achieving the
highest average quality score is retained as the
task-specific prior $m_d^{*}$. Repeating this procedure
across restoration tasks produces a compact task-prior
bank, enabling multiple callable restoration actions
from a single frozen backbone. 
% The detailed scoring
% formulation and search settings are provided in the
% supplementary material.

\subsection{Sequential Policy Learning}
\label{sec:policy_learning}
% Given the constructed shared-backbone toolbox, we train a state-conditioned policy \(\pi_\phi\) to select restoration actions according to the evolving image state.
% As illustrated in Fig.~\ref{pipeline}(c), the policy is initialized through supervised fine-tuning (SFT) using restoration state–action pairs obtained from offline trajectory exploration.
% We then employ Group Relative Policy Optimization (GRPO) to refine action selection using quality-grouped process-aware rewards.

Given the constructed shared-backbone toolbox, we learn
a state-conditioned policy $\pi_\phi$ to sequentially
select restoration actions according to the evolving
image state. As illustrated in Fig.~\ref{pipeline}(c), we first
initialize the policy through supervised fine-tuning
(SFT), followed by Group Relative Policy Optimization
(GRPO) using quality-grouped process-aware rewards.

% Given the constructed shared-backbone toolbox, we learn
% a state-conditioned policy $\pi_\phi$ to sequentially
% select restoration actions according to the evolving
% image state.
% As illustrated in Fig.~\ref{pipeline}(c),
% we first initialize the policy through supervised
% fine-tuning (SFT) and construct quality-grouped process-aware rewards to evaluate restoration trajectories.
% The resulting reward signals are then incorporated
% into Group Relative Policy Optimization (GRPO)
% to refine the policy.
\noindent\textbf{Supervised Fine-Tuning.}
To provide a reliable initialization for policy
optimization, we construct an offline supervision
dataset $\mathcal{D}_{\mathrm{SFT}}=\{(I_t,a_t^*)\}$,
where each intermediate restoration state is paired
with an action selected according to restoration-quality
improvement during offline trajectory exploration.
The policy is initialized by minimizing the classification loss
\begin{equation}
\mathcal L_{\mathrm{SFT}}
=
-\mathbb E_{(I_t,a_t^*)\sim\mathcal D_{\mathrm{SFT}}}
\left[\log\pi_\phi(a_t^*\mid I_t)\right].
\label{eq:sft}
\end{equation}
A frozen copy is retained as the reference policy
$\pi_{\mathrm{ref}}$ for subsequent GRPO optimization.

% \noindent\textbf{Supervised fine-tuning.}
% Directly optimizing the policy through reinforcement
% learning may lead to inefficient exploration of restoration
% sequences. We therefore construct an offline supervision
% dataset $\mathcal D_{\mathrm{SFT}}=\{(I_t,a_t^*)\}$,
% where each intermediate restoration state is paired with
% an action selected from candidate trajectories according
% to restoration-quality improvement.
% The policy is initialized by minimizing the classification loss
% \begin{equation}
% \mathcal L_{\mathrm{SFT}}
% =
% -\mathbb E_{(I_t,a_t^*)\sim\mathcal D_{\mathrm{SFT}}}
% \left[\log\pi_\phi(a_t^*\mid I_t)\right].
% \label{eq:sft}
% \end{equation}
% After SFT, the initialized policy serves as the starting
% point for reinforcement learning, while a frozen copy
% is retained as the reference policy $\pi_{\mathrm{ref}}$
% to constrain policy updates during GRPO optimization.

\noindent\textbf{Quality-Grouped Process-Aware Reward.}
To evaluate restoration outcomes, we organize
quality metrics into two groups: fidelity $F$ (PSNR, SSIM and LPIPS) and perceptual quality $Q$ (CLIP-IQA, MUSIQ and MANIQA).
The two groups are separately standardized across trajectories sampled for the
same input and combined as
\begin{equation}
D_G(F,Q)=0.45N_G(F)+0.55N_G(Q),
\end{equation}
where $N_G(\cdot)$ denotes standardization within
the trajectory group $G$. 
This formulation reduces
the influence of scale differences between fidelity
and perceptual feedback.
To incorporate intermediate restoration outcomes,
we compute the stepwise quality changes
$\Delta F_{i,t}$ and $\Delta Q_{i,t}$ for trajectory $i$.
% To incorporate intermediate restoration outcomes,
% we compute the stepwise quality changes
% $\Delta F_{i,t}$ and
% $\Delta Q_{i,t}$.
The process reward and final advantage are defined as

\begin{equation}
\begin{aligned}
r_i^{\mathrm{term}} &= D_G(F_{i,T},Q_{i,T}),\\
r_{i,t}^{\mathrm{proc}} &= D_{G_t}(\Delta F_{i,t},\Delta Q_{i,t}),\\
A_{i,t} &= r_i^{\mathrm{term}}
+\sum_{k=t+1}^{T}r_{i,k}^{\mathrm{proc}},
\end{aligned}
\end{equation}
% \begin{equation}
% \begin{aligned}
% r_{i,t}^{\mathrm{proc}}
% &=D_{G_t}(\Delta F_{i,t},\Delta Q_{i,t}),\\
% A_{i,t}
% &=D_G(F_{i,T},Q_{i,T})
% +\sum_{k=t}^{T}r_{i,k}^{\mathrm{proc}},
% \end{aligned}
% % \vspace{-1mm}
% \end{equation}
where $G_t$ denotes the trajectories reaching step $t$.
The resulting advantage combines terminal quality
evaluation with return-to-go from stepwise quality
changes, providing step-dependent feedback for
policy optimization. 
% Detailed metric aggregation
% is provided in the supplementary material.
\begin{figure*}[!htb]
    \centering
    \includegraphics[width=\textwidth]{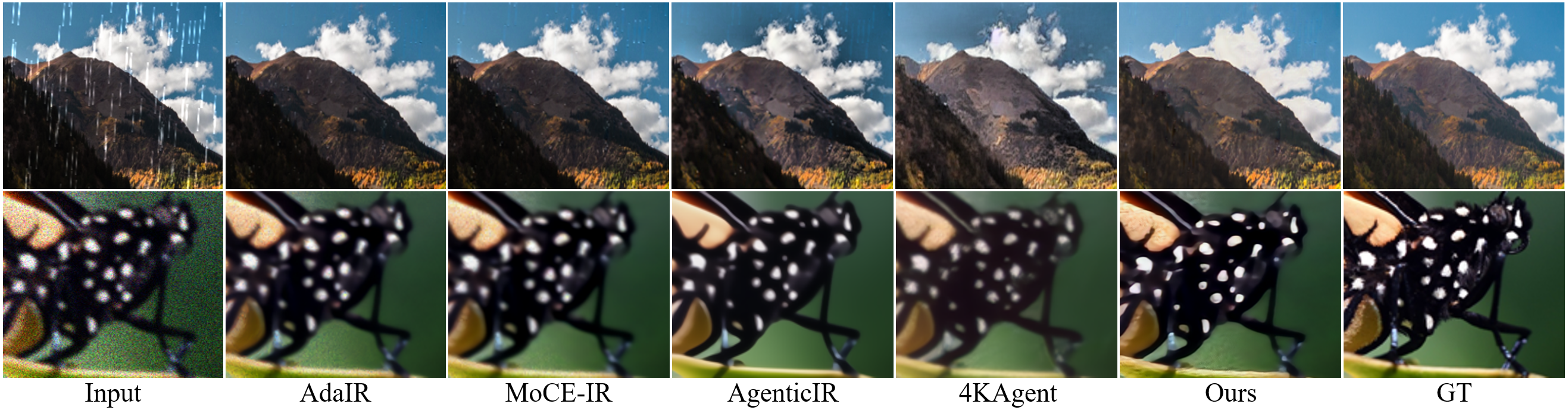}
    \vspace{-6mm}
        \caption{Qualitative comparison on synthetic composite degradations.
    Top: low light and rain.
    Bottom: blur and noise.}
%     \caption{Qualitative comparisons on synthetic
% composite-degradation images.
% The top row shows a mountain scene with low light
% and rain, while the bottom row shows a butterfly
% affected by defocus blur, motion blur, and noise.
% Our method suppresses these degradations while
% preserving structural details and natural appearance.}

    \label{fig:visual_comparison1}
    \vspace{-4mm}
\end{figure*}

\noindent\textbf{GRPO Optimization.}
Starting from the SFT-initialized policy, we employ
Group Relative Policy Optimization (GRPO) to
refine sequential action selection
using the stepwise advantages defined above.
For each input image, the rollout policy
$\pi_{\mathrm{old}}$ samples a group of restoration
trajectories.
The clipped surrogate objective is defined as
\begin{equation}
\begin{aligned}
\ell_{i,t}(\phi)=\min\big\{&
\rho_{i,t}(\phi)A_{i,t},\\
&\operatorname{clip}\big(
\rho_{i,t}(\phi),1-\epsilon,1+\epsilon
\big)A_{i,t}
\big\}.
\end{aligned}
\end{equation}
The overall objective includes a KL constraint
to the reference policy:
\begin{equation}
\begin{aligned}
\mathcal{L}_{\mathrm{GRPO}}
=&-\mathbb{E}_{i,t}\big[\ell_{i,t}(\phi)\big]\\
&+\beta\mathbb{E}_{i,t}\big[
D_{\mathrm{KL}}(\pi_\phi\|\pi_{\mathrm{ref}})
\big].
\end{aligned}
\end{equation}
Here, $\rho_{i,t}(\phi)$ is the
probability ratio between the current and rollout
policies for the selected action.
The clipping threshold and KL coefficient are set
to $\epsilon=0.2$ and $\beta=0.03$, respectively.
After policy optimization, we freeze the learned
policy and perform a lightweight refinement of
selected task-specific modulation coefficients,
keeping the shared restoration backbone unchanged.

%% file: sec/4_experiment.tex
\section{Experiments}
% We conduct extensive experiments to comprehensively evaluate the restoration performance, tool discriminability, sequential decision-making capability, and computational efficiency of our method. 
% Specifically, we compare our method with \(\mathcal{N}_{A}\) state-of-the-art all-in-one restoration methods and \(\mathcal{N}_{G}\) agent-based methods on three paired synthetic composite-degradation benchmarks and a paired real-world multi-task restoration benchmark.
% We further conduct comprehensive ablation studies and diagnostic analyses to investigate the construction of the shared-parameter toolbox, the effectiveness of policy learning strategies, the proposed factorized process-aware reward, and the resulting restoration trajectories. 
% Finally, we analyze model complexity and inference efficiency to assess the practical advantages of using a single shared restoration backbone.
% We conduct extensive experiments to evaluate the restoration performance, sequential decision-making capability, and computational efficiency of our method. 
% We compare our approach with 8 all-in-one methods and 6 agent-based methods on three synthetic composite-degradation benchmarks and a real-world multi-task benchmark. 
% We further analyze the shared-parameter toolbox, policy learning strategies, reward design, restoration trajectories, and model complexity.
\subsection{Experimental Setup}
% \subsubsection{Datasets}
\noindent\textbf{Datasets.}
Following MiOIR~\cite{kong2024mioir}, we use DIV2K~\cite{agustsson2017div2k}
and Flickr2K~\cite{lim2017edsr} as source images to synthesize
approximately 25,000 paired training samples covering
48 composite-degradation combinations across eight
restoration tasks. 
For toolbox construction and evaluation, we
generate eight single-degradation datasets,
each containing 200 paired images for prior
search and 100 disjoint pairs for testing.
% For toolbox construction, we
% additionally generate eight single-degradation subsets,
% each containing 200 paired images for task-prior search.
Evaluation is conducted on the MiO100-based Groups A,
B, and C following AgenticIR~\cite{zhu2025agenticir}, comprising
640, 400, and 400 images, respectively, with two-
or three-degradation combinations.
% We further evaluate on real-world degraded images
% collected from SIDD~\cite{abdelhamed2018sidd}, LHP-Rain~\cite{guo2023lhprain},
% and NH-HAZE~\cite{ancuti2020nhhaze}.
All training and prior-search images
are selected to avoid overlap with the source
images used in the MiO100 evaluation sets.
We further evaluate on a real-world multi-task benchmark comprising 280 degraded images selected from LHP-Rain~\cite{guo2023lhprain} and RTTS~\cite{li2019benchmarking}.
\noindent\textbf{Evaluation Metrics.}
% Following AgenticIR~\cite{zhu2025agenticir}, we evaluate the synthetic paired benchmarks using six metrics. PSNR, SSIM~\cite{wang2004ssim}, and LPIPS~\cite{zhang2018lpips} assess full-reference restoration quality, while MANIQA~\cite{yang2022maniqa}, CLIP-IQA~\cite{wang2023clipiqa}, and MUSIQ~\cite{ke2021musiq} measure no-reference perceptual quality. 
% For the real-world benchmark, we report four no-reference metrics:NIQE~\cite{mittal2013niqe}, MANIQA, CLIP-IQA, and MUSIQ. Higher values indicate better performance for all metrics except LPIPS and NIQE. 
We report PSNR, SSIM~\cite{wang2004ssim}, LPIPS~\cite{zhang2018lpips},
MANIQA~\cite{yang2022maniqa}, CLIP-IQA~\cite{wang2023clipiqa}, and
MUSIQ~\cite{ke2021musiq} on synthetic benchmarks.
For real-world evaluation, we use four no-reference
metrics: NIQE~\cite{mittal2013niqe}, MANIQA, CLIP-IQA, and MUSIQ.
% Higher scores indicate better performance except
% for LPIPS and NIQE.

\noindent\textbf{Implementation Details.}
% Our framework is implemented in PyTorch and trained on four NVIDIA RTX 3090 GPUs.
% We adopt AdaIR~\cite{cui2025adair} as the shared restoration backbone and keep its parameters frozen throughout offline toolbox construction and policy learning.
% The policy network adopts an ImageNet-pretrained~\cite{deng2009imagenet} ResNet-34~\cite{he2016resnet} with an eight-way classification head and is initialized by supervised fine-tuning before GRPO training. 
% The policy is first trained for 30 epochs using an initial learning rate of \(1\times10^{-4}\).
% During GRPO, the restoration backbone and searched task priors are frozen, while only the policy network is optimized.
% For each input image, we sample a group of 8 trajectories with a maximum length of 6 restoration steps. 
% The policy is trained for 50 epochs using AdamW~\cite{loshchilov2019adamw} with an initial learning rate of \(3\times10^{-5}\), which is decayed to 10\% of its initial value following a cosine schedule.
% After GRPO optimization, we freeze the learned policy and fine-tune only a small subset of task-specific modulation parameters for 7 epochs, while keeping the shared AdaIR backbone fixed.
% Training images are cropped to \(256\times256\), and the global batch size is 8.
We adopt AdaIR~\cite{cui2025adair} as the frozen restoration
backbone and an ImageNet-pretrained ResNet34~\cite{he2016resnet}
with an eight-way classification head as the policy network.
For offline prior search, we set $\rho=0.5$ and $K=5$. We perform two search sweeps over
the candidate coefficients
$\{0, 0.25, 0.5, 0.75, 1.0, 1.1, 1.2\}$.
The policy undergoes SFT for 30 epochs with an initial
learning rate of $1\times10^{-4}$, followed by GRPO
for 50 epochs using AdamW~\cite{loshchilov2019adamw} with an initial
learning rate of $3\times10^{-5}$ and cosine decay to
10\% of its initial value.
For each GRPO input, we sample 8 trajectories,
each executing exactly 6 restoration actions
without early stopping.
% Each GRPO input generates 8 trajectories of up to
% 6 restoration steps.
We then freeze the policy and refine selected modulation
parameters for 7 epochs while keeping the backbone
fixed. Training uses $256\times256$ image crops,
a global batch size of 8, and 4 NVIDIA RTX
3090 GPUs.
\subsection{Comparison with State-of-the-Art Methods}
\begin{table}[t]
\centering
\caption{Quantitative comparison on three multiple-degradation subsets
of MiO100. The best, second-best, and third-best results are marked in
\textbf{bold}, \underline{underline}, and \textit{italic}, respectively.}
\vspace{-2mm}
\label{tab:sota_mio100}
\setlength{\tabcolsep}{1.5pt}       % 列间距
\setlength{\aboverulesep}{0.25ex}   % 横线上方间距
\setlength{\belowrulesep}{0.25ex}   % 横线下方间距
\resizebox{\columnwidth}{!}{%
% \begin{tabular}{
%     @{}
%     c@{\hspace{2pt}}       % Type → Method
%     l@{\hspace{4pt}}       % Method → PSNR
%     *{7}{c}
%     @{}
% }
\begin{tabular}{l c c c c c c }
\toprule
\textbf{Method}&
\textbf{~~PSNR~$\uparrow$} & \textbf{~~SSIM~$\uparrow$} & \textbf{~~LPIPS~$\downarrow$} &
\textbf{MANIQA~$\uparrow$} & \textbf{CLIP-IQA~$\uparrow$} & \textbf{MUSIQ~$\uparrow$}\\
% Method&
% PSNR & SSIM & LPIPS$\downarrow$ &
% MANIQA & CLIP-IQA& MUSIQ  \\
\midrule
% \midrule
\rowcolor{black!5}
\multicolumn{7}{c}{\textbf{Group A}} \\
\midrule
% ==================== Group A ====================
% \multirow{12}{*}{Group A}
% \multirow{8}{*}{All-in-One}
% &&&Group A&&&
AirNet~\cite{li2022airnet}
 &19.13 & 0.6019 & 0.4283 & 0.2581 & 0.3930 & 42.46  \\
PromptIR~\cite{potlapalli2023promptir}
& 20.06 & 0.6088 & 0.4127 & 0.2633 & 0.4013 & 42.62 \\
MiOIR~\cite{kong2024mioir}
&20.84 & 0.6558 & 0.3715 & 0.2451 & 0.3933 & 47.82  \\
DA-CLIP~\cite{luo2024daclip}
 &19.58 & 0.6032 & 0.4266 & 0.2418 & 0.4139 & 42.51  \\
InstructIR~\cite{conde2024instructir}
 &18.03 & 0.5751 & 0.4429 & 0.2660 & 0.3528 & 45.77 \\
AutoDIR~\cite{jiang2024autodir}
&19.64 & 0.6286 & 0.3967 & 0.2500 & 0.3767 & 47.01 \\
AdaIR~\cite{cui2025adair}
& 19.07 & 0.6063 & 0.4162 & 0.2629 & 0.3944 & 44.71\\
MoCE-IR~\cite{zamfir2025moceir}
 &18.75 & 0.6010 & 0.4218 & 0.2653 & 0.4037 & 44.27 \\
\cmidrule(lr){1-7}
% \multirow{6}{*}{Agent}
AgenticIR~\cite{zhu2025agenticir}
& 21.04 & 0.6818 & 0.3148 & 0.3071 & 0.4474 & 56.88 \\
MAIR~\cite{jiang2026mair}
 &21.02 & 0.6715 & \third{0.2963}
&0.3330 & 0.4751 & 59.19 \\
4KAgent~\cite{zuo20254kagent}
& \second{21.48} & 0.6720 & 0.3019
&\third{0.3748} & \second{0.5544} & \second{63.19} \\
TIR-Agent~\cite{jia2026tiragent} & \best{22.07} & \second{0.6935} &  \second{0.2874} & \second{0.3907} & \third{0.5454} & \best{63.69} \\
SEAR~\cite{cui2026sear} & 20.11 & 0.6503 & 0.3018 & 0.3430 & 0.5039 & \third{61.29}  \\
RAR~\cite{chen2026rar} & 20.46 & \best{0.7144} & \best{0.1299} & \best{0.4659} & \best{0.6566} & 57.19 \\
\rowcolor{gray!20}
 PrismRestore (Ours)
& \third{21.19} & \third{0.6826} & 0.3221
& 0.3104 & 0.4445 & 54.24 \\

% \midrule
\midrule
\rowcolor{black!5}
\multicolumn{7}{c}{\textbf{Group B}} \\
\midrule

% ==================== Group B ====================
% \multirow{12}{*}{Group B}
% & \multirow{8}{*}{All-in-One}
 AirNet~\cite{li2022airnet}
& 19.31 & 0.6567 & 0.3670 & 0.2882 & 0.4274 & 47.88  \\
 PromptIR~\cite{potlapalli2023promptir}
& 20.47 & 0.6704 & 0.3370 & 0.2893 & 0.4289 & 48.10\\
 MiOIR~\cite{kong2024mioir}
& 20.56 & 0.6905 & 0.3243 & 0.2638 & 0.4330 & 51.87 \\
 DA-CLIP~\cite{luo2024daclip}
& 18.56 & 0.5946 & 0.4405 & 0.2435 & 0.4154 & 43.70 \\
 InstructIR~\cite{conde2024instructir}
& 18.34 & 0.6235 & 0.4072 & 0.3022 & 0.3790 & 50.94 \\
 AutoDIR~\cite{jiang2024autodir}
& 19.90 & 0.6643 & 0.3542 & 0.2534 & 0.3986 & 49.64 \\
 AdaIR~\cite{cui2025adair}
& 19.25 & 0.6638 & 0.3456 & 0.2899 & 0.4198  & 49.28  \\
 MoCE-IR~\cite{zamfir2025moceir}
& 18.66 & 0.6533 & 0.3540 & 0.2899 & 0.4320 & 48.55\\
\cmidrule(lr){1-7}
% & \multirow{6}{*}{Agent}
 AgenticIR~\cite{zhu2025agenticir}
& 20.55 & 0.7009 & 0.3072 & 0.3204 & 0.4648 & 57.57  \\
 MAIR~\cite{jiang2026mair}
& 20.92 & 0.7004 & 0.2788
& 0.3544 & 0.5084 & \third{60.98}  \\
 4KAgent~\cite{zuo20254kagent}
& 20.95 & 0.6727 & 0.3017
& \third{0.3734} & \second{0.5505} & \second{62.69} \\
TIR-Agent~\cite{jia2026tiragent} & \best{22.80} & \second{0.7340} & \third{0.2725} & \second{0.3997} & \third{0.5488} & \best{65.24}  \\
SEAR~\cite{cui2026sear} & 20.48 & 0.6824 & 0.2889 & 0.3464 & 0.5134 & 60.65 \\
RAR~\cite{chen2026rar} & \third{21.04} & \third{0.7326} & \best{0.1269} & \best{0.4582} &\best{0.6483} & 56.91 \\
\rowcolor{gray!20}
  PrismRestore (Ours)
& \second{22.48} & \best{0.7456} & \second{0.2676}
& 0.3248 & 0.4507 & 57.42 \\

% \midrule
\midrule
\rowcolor{black!5}
\multicolumn{7}{c}{\textbf{Group C}} \\
\midrule
% ==================== Group C ====================
% \multirow{12}{*}{Group C}
% & \multirow{8}{*}{All-in-One}
 AirNet~\cite{li2022airnet}
& 17.95 & 0.5145 & 0.5782 & 0.1854 & 0.3113 & 30.12 \\
 PromptIR~\cite{potlapalli2023promptir}
& 18.51 & 0.5166 & 0.5756 & 0.1906 & 0.3104 & 29.71 \\
 MiOIR~\cite{kong2024mioir}
& 15.63 & 0.4896 & 0.5376 & 0.1717 & 0.2891 & 37.95  \\
 DA-CLIP~\cite{luo2024daclip}
& 18.53 & 0.5320 & 0.5335 & 0.1916 & 0.3476 & 33.87  \\
 InstructIR~\cite{conde2024instructir}
& 17.09 & 0.5135 & 0.5582 & 0.1732 & 0.2537 & 33.69 \\
 AutoDIR~\cite{jiang2024autodir}
& 18.61 & 0.5443 & 0.5019 & 0.2045 & 0.2939 & 37.86  \\
 AdaIR~\cite{cui2025adair}
& 17.98 & 0.5221 & 0.5787 & 0.1895 & 0.3085 & 30.95 \\
 MoCE-IR~\cite{zamfir2025moceir}
& 17.85 & 0.5302 & 0.5876 & 0.1862 & 0.3124 & 29.82 \\
\cmidrule(lr){1-7}
% & \multirow{6}{*}{Agent}
 AgenticIR~\cite{zhu2025agenticir}
& 18.82 & 0.5474 & 0.4493 & 0.2698 & 0.3948 & 48.68\\
 MAIR~\cite{jiang2026mair}
& 19.42 & 0.5544 & \third{0.4142}
& 0.2798 & 0.4239 & 51.36  \\
 4KAgent~\cite{zuo20254kagent}
& \best{19.77} & 0.5629 & 0.4271
& \third{0.3545} & \third{0.5233} & \third{55.56} \\
TIR-Agent~\cite{jia2026tiragent} & \third{19.53} & \third{0.5643} & \second{0.4120} & \second{0.3739} & \second{0.5495} & \best{56.93} \\
SEAR~\cite{cui2026sear} & 18.99 & 0.5402 & 0.4173 & 0.2733 & 0.4419 & 51.73  \\
RAR~\cite{chen2026rar} & 19.33 & \best{0.6579} & \best{0.1489} & \best{0.4653} & \best{0.6554} & \second{56.56}  \\

\rowcolor{gray!20}
 PrismRestore (Ours)
& \second{19.74} & \second{0.5726} & 0.4488
& 0.2260 & 0.3552 & 45.14 \\

\bottomrule
\end{tabular}%
}
\vspace{-4mm}
\end{table}

\noindent\textbf{Synthetic Composite-Degradation Benchmarks.}
% As shown in Tab.~\ref{tab:sota_mio100}, our method
% achieves an average PSNR of 21.14 dB and MUSIQ
% of 52.27 across Groups A, B, and C, exceeding the
% strongest all-in-one baseline in each group by
% 1.13 dB and 6.39 points on average, respectively.
% % Among the evaluated methods, it ranks second in
% % PSNR on Groups B and C, achieving 22.48 and
% % 19.74 dB, respectively.
% Compared with recent agentic methods, our approach remains competitive, ranking second in PSNR on Groups B and C with 22.48 and 19.74 dB, respectively, and achieving the highest SSIM of 0.7456 on Group B. Although methods such as TIR-Agent and RAR attain stronger results on certain fidelity or perceptual metrics, our method achieves competitive restoration quality using restoration actions derived from a single frozen backbone.
On MiO100 (Tab.~\ref{tab:sota_mio100}), PrismRestore outperforms the eight listed all-in-one baselines on all six metrics in each group.
Averaged equally over the three groups, the gains over the best all-in-one result in each group are 1.13~dB in PSNR and 6.39 points in MUSIQ.
Despite using a single restoration backbone,
PrismRestore ranks second in PSNR on Groups B
and C and achieves the highest SSIM (0.7456)
and second-lowest LPIPS (0.2676) on Group B.
% Among all compared methods, 
% PrismRestore ranks second in PSNR on Groups B and C, and achieves the highest SSIM (0.7456) and second-lowest LPIPS (0.2676) on Group B.
TIR-Agent obtains higher no-reference scores in all three groups, while RAR achieves lower LPIPS.
Fig.~\ref{fig:visual_comparison1} shows two composite-degradation examples.
For low light and rain, PrismRestore suppresses rain streaks and improves mountain visibility.
For the combination of defocus blur, motion blur, and noise, it preserves the butterfly's wing patterns and fine boundaries.

\begin{figure*}[t]
    \centering
    \includegraphics[width=\textwidth]{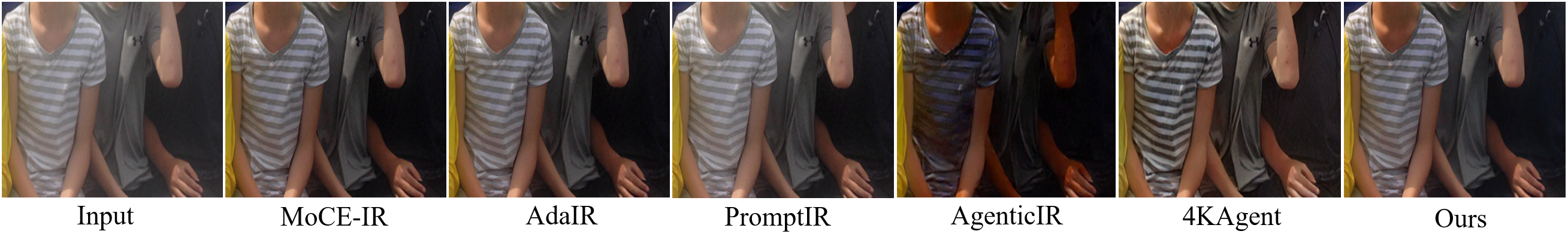}
    \vspace{-6mm}
    \caption{Qualitative comparison on a real-world hazy scene.
Our method effectively removes haze while preserving
fine details with fewer visible artifacts.
Please zoom in for a better view.}
    \vspace{-4mm}
    \label{fig:real_world}
\end{figure*}

\noindent\textbf{Real-World Multi-Task Restoration Benchmark.}
% We further evaluate our method on the real-world multi-task benchmark.
% As reported in Tab.~\ref{tab:real_world_unpaired}, our method achieves the lowest
% NIQE of 5.70 among the evaluated methods and outperforms
% all compared all-in-one baselines across the four
% no-reference metrics.
% Although AgenticIR achieves higher MANIQA, CLIP-IQA, and MUSIQ scores, indicating that a gap remains in perceptual quality, our method maintains competitive perceptual quality with substantially fewer restoration-toolbox parameters.
% On the unpaired real-world benchmark (Tab.~\ref{tab:real_world_unpaired}), PrismRestore achieves the lowest NIQE of 5.70, compared with 6.20 for the best all-in-one baseline and 6.51 for AgenticIR.
% It also exceeds the listed all-in-one baselines on CLIP-IQA and MUSIQ, while its MANIQA is close to AutoDIR (0.3470 vs.\ 0.3469).
% AgenticIR obtains higher MANIQA, CLIP-IQA, and MUSIQ scores.
On the real-world benchmark (Tab.~\ref{tab:real_world}),
PrismRestore outperforms all six all-in-one baselines on all four
metrics. Among the evaluated methods, it achieves the highest MANIQA
(0.3721) and CLIP-IQA (0.4216) and ranks second in NIQE (5.07).
It also outperforms AgenticIR on NIQE, MANIQA, and CLIP-IQA,
with a slightly lower MUSIQ score (62.08 vs.\ 62.29).
4KAgent achieves the best NIQE (4.77) and MUSIQ (63.66).
As shown in Fig.~\ref{fig:real_world}, the all-in-one
methods exhibit more residual haze, while 4KAgent
retains noticeable noise in local regions.
AgenticIR produces overly enhanced textures and
unnatural color variations. In contrast,
PrismRestore achieves more effective haze removal
with cleaner details and more natural visual
appearance.
% The pedestrian example in Fig.~\ref{fig:real_world} shows improved visibility with clothing contours preserved.
% Although AgenticIR achieves higher MANIQA, CLIP-IQA,
% and MUSIQ scores, this gap may be partly attributed
% to the limited capacity of our single-backbone toolbox
% compared with its collection of independently
% parameterized restoration models.
% Nevertheless, our method maintains competitive
% perceptual quality with substantially fewer
% restoration parameters,
% as further analyzed
% in Sec.~\ref{sec:complexity}.
% As illustrated in Fig.~\ref{fig:real_world},
% our method improves pedestrian visibility while
% preserving clothing contours and structural details.

\begin{table}[t]
\centering
% \vspace{-2mm}
\caption{Quantitative comparison on the real-world multi-task benchmark.}
\vspace{-2mm}
% \begin{adjustbox}{width=\linewidth}
\label{tab:real_world}

\fontsize{8.5pt}{10pt}\selectfont
\setlength{\tabcolsep}{4pt}
\resizebox{\columnwidth}{!}{
\begin{tabular}{lcccc}
\toprule
\textbf{Method}
& \textbf{NIQE~$\downarrow$}
& \textbf{MANIQA~$\uparrow$}
& \textbf{CLIP-IQA~$\uparrow$}
& \textbf{MUSIQ~$\uparrow$}\\
\midrule
AirNet
& 5.20 & 0.3484 & 0.3479 & 58.64 \\
PromptIR
& 5.50 & 0.3501 & 0.3566 & 58.89 \\
AdaIR
& 5.42 & \second{0.3643} & \third{0.3808}  &59.52  \\
MoCE-IR
& 6.64
& 0.3348
&  0.3353
&  51.02\\
AutoDIR
& 5.52
& 0.3590
& 0.3671
& 59.43\\
InstructIR
& 5.40
& 0.3600
& 0.3695
& 59.01\\
\midrule
4KAgent
& \best{4.77}
& \third{0.3604}
& \second{0.3997}
& \best{63.66} \\
AgenticIR
& \third{5.13}
& 0.3479
& 0.3751
& \second{62.29} \\
\rowcolor{gray!15}
 PrismRestore (Ours)
& \second{5.07}
&\best{0.3721}
& \best{0.4216}
& \third{62.08} \\
\bottomrule
\end{tabular}}
% \end{adjustbox}
\vspace{-2mm}
\end{table}

\noindent\textbf{Toolbox Complexity and Inference Efficiency.}
PrismRestore provides eight restoration actions from a single frozen backbone, using 28.78M restoration-toolbox parameters (Tabs.~\ref{tab:complexity} and~\ref{tab:time_psnr}).
Relative to HybridAgent, which also shares a restoration backbone, this represents a 13.9-fold reduction in toolbox parameters.
The corresponding reductions relative to AgenticIR and 4KAgent are approximately 124-fold and 216-fold, respectively.
Action selection uses a ResNet34 policy, with an average of 4.65 tool calls per image, compared with 8.31 for AgenticIR and 13.23 for 4KAgent.
PrismRestore achieves a mean PSNR of 21.14~dB across the three groups with an average inference time of 5.00~s per image.
Compared with TIR-Agent, it is 15.4$\times$ faster with 0.33~dB lower mean PSNR.
Compared with RAR, it achieves 0.86~dB higher mean PSNR and reduces inference time from 8.45 to 5.00~s.

\begin{table}[t]
\centering
\caption{Toolbox complexity and inference overhead. 
% Tools reports the number of restoration actions,
% with distinct backbone counts in parentheses;
% Params.T/P denote toolbox/planner parameters.
Tools denotes actions (distinct backbones);
Params.T/P denote toolbox/planner parameters.
Calls are averaged over three groups of MiO100.
HybridAgent reports FastAgent parameters only.}
 \vspace{-2mm}
\label{tab:complexity}
\fontsize{8.5pt}{10pt}\selectfont
% \footnotesize
    \setlength{\tabcolsep}{3.2pt}
\resizebox{\columnwidth}{!}{
\begin{tabular}{lccccc}
\toprule
\textbf{Method} & \textbf{Tools }&\textbf{ Params.T}  & \textbf{Planner}&\textbf{ Params.P}& \textbf{Calls}  \\
\midrule
AgenticIR & 33 & 3579.67M & GPT-4& $>$175B (GPT-3)& 8.31  \\

4KAgent   & 64 & 6205.92M &GPT-4& $>$175B (GPT-3)&  13.23  \\

HybridAgent& 11(1) & 400.28M &  Llama3.2 & 1.23B& -- \\
% \midrule
\rowcolor{gray!15}
 PrismRestore (Ours)      & 8(1)   & \best{28.78M}  & ResNet34& \best{0.02B}& \best{4.65} \\
\bottomrule
\end{tabular}}\vspace{-2mm}
\end{table}

% \noindent\textbf{Toolbox Complexity and Inference Efficiency.}
% % As shown in Tab.~\ref{tab:complexity}, our framework provides eight
% % restoration actions using a single shared backbone
% % with only 28.78M restoration parameters, approximately
% % 124$\times$ and 216$\times$ fewer than AgenticIR and
% % 4KAgent, respectively. Moreover, our method requires
% % only 4.65 tool calls per image, compared with 8.31
% % and 13.23 for AgenticIR and 4KAgent, respectively.
% % Together with the quality--efficiency comparisons in
% % Tab.~2, these results demonstrate that our framework
% % maintains competitive restoration performance while
% % substantially reducing toolbox storage and tool
% % invocation overhead.

% As shown in Tab.~\ref{tab:complexity}, our framework provides eight
% restoration actions using a single shared backbone
% with only 28.78M restoration parameters,
% approximately 124$\times$ and 216$\times$ fewer
% than AgenticIR and 4KAgent, respectively.
% Moreover, our method requires only 4.65 tool calls
% per image, compared with 8.31 and 13.23 for
% AgenticIR and 4KAgent, respectively.
% Together with the restoration
% quality and inference-time comparisons in Tab.~\ref{tab:time_psnr},
% these results highlight the efficiency advantages
% of constructing multiple restoration actions from
% a shared backbone, reducing both toolbox storage
% and online tool invocation overhead.

\begin{table}[t]
    \centering
    \caption{Average restoration quality and inference time. The best, second-best, and third-best results are marked in bold, underline, and italic, respectively.}
\vspace{-2mm}
    \label{tab:time_psnr}

    \fontsize{8.5pt}{10pt}\selectfont
    \setlength{\tabcolsep}{2.6pt}
    \renewcommand{\arraystretch}{0.95}
    \setlength{\aboverulesep}{0.25ex}
    \setlength{\belowrulesep}{0.25ex}
\resizebox{\columnwidth}{!}{
    % \begin{adjustbox}{width=\linewidth}
    \begin{tabular}{l ccccc}
        \toprule
        \rowcolor{black!5}
        \multicolumn{6}{c}{\textbf{All-in-one}} \\
        \midrule

        \textbf{Method}
        & AirNet & PromptIR & MiOIR & AutoDIR & AdaIR \\
        \midrule

        \textbf{PSNR}
        & 18.80 & 19.68 & 19.01 & 19.38 & 18.77 \\

        \textbf{Time(s)}
        & 1.88 & 1.31 & 1.28 & 181.15 & 1.35 \\

        \midrule
        \rowcolor{black!5}
        \multicolumn{6}{c}{\textbf{Agent}} \\
        \midrule

        \textbf{Method}
        & AgenticIR & 4KAgent & TIR-Agent & RAR
        &  PrismRestore (Ours) \\
        \midrule

        \textbf{PSNR}
        & 20.14
        & \third{20.73}
        & \best{21.47}
        & 20.28
        & \second{21.14} \\

        \textbf{Time(s)}
        & 211.84
        & 432.58
        & \third{77.20}
        & \second{8.45}
        & \best{5.00} \\

        \bottomrule
    \end{tabular}}
    \vspace{-2mm}
    % \end{adjustbox}
    % \vspace{-2mm}
\end{table}

\subsection{Analysis of the Shared-Backbone Toolbox}

% \subsubsection{Ablation of Offline Toolbox Construction}
% \noindent\textbf{Ablation of Offline Toolbox Construction.}
\noindent\textbf{Component Ablation.}
Tab.~\ref{tab:toolbox_component} reports mean PSNR improvement over the unmodulated AdaIR backbone (Gain) and image-level task--action matching (Match).
Random partitioning achieves 1.17~dB Gain and 38.1\% Match.
With representative partitioning retained, activation-only and gradient-only variants achieve 2.16~dB/58.3\% and 2.23~dB/59.1\%, respectively.
Combining both statistics increases Gain to 2.69~dB and Match to 71.6\%.

Restricting modulation to non-representative channels increases Match from 64.6\% to 71.6\% relative to unrestricted modulation, with a 0.12~dB increase in Gain.
Searching representative channels instead yields 2.01~dB Gain and 52.7\% Match.
The comparison supports retaining representative channels when constructing actions with stronger task preferences.

\begin{figure}[t]
    \centering
    \includegraphics[width=0.90\columnwidth]{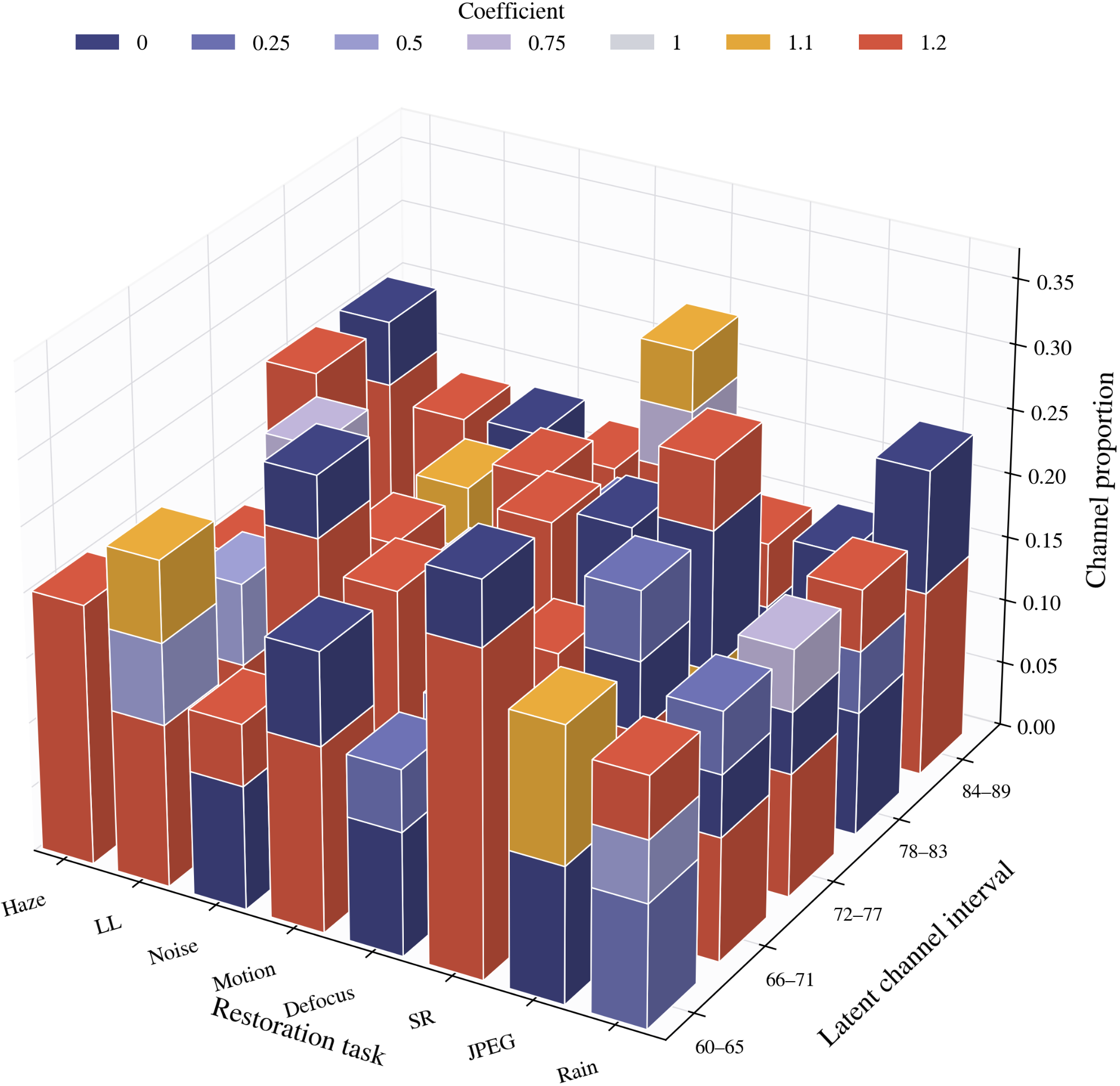}
    \caption{Task-specific modulation patterns of the searched
priors in the latent stage. Bar heights indicate the
proportion of modulated channels across five selected
intervals, while colors represent modulation coefficients
stacked by frequency.}
    \vspace{-3mm}
    \label{fig:prior_distribution}
    \vspace{-2mm}
\end{figure}

\begin{table}[t]

\caption{Ablation of shared-backbone toolbox construction.
Act., Grad., and Part. denote activation, gradient,
and channel partitioning, respectively.
Rep. Search uses representative channels.
Gain and Match denote PSNR improvement
and Top-1 task--action matching accuracy.}
% \vspace{-2mm}
% \begin{adjustbox}{width=\linewidth}
\centering

\label{tab:toolbox_component}
\fontsize{8.5pt}{10pt}\selectfont
\setlength{\tabcolsep}{2.6pt}
% \renewcommand{\arraystretch}{1.08}
% \resizebox{\columnwidth}{!}{%
\begin{tabular}{lccccc}
\toprule
\textbf{Setting} & \textbf{Act.} &\textbf{ Grad.} &\textbf{Part.}
& \textbf{Gain(dB)}
& \textbf{Match(\%)} \\
\midrule
% Learned
% & -- & -- & -- & --
% % & 1.53 & 53.9 \\
% Random Partition
% & & & \checkmark & 1.17 & 38.1 \\
Random Part.
&   &  &\checkmark & 1.17 & 38.1 \\
w/o Grad.
& \checkmark &  & \checkmark & 2.16 & 58.3 \\
w/o Act.
& & \checkmark  & \checkmark & 2.23 & 59.1 \\
w/o Part.
& \checkmark  & \checkmark & & \second{2.57} & \second{64.6} \\
Rep. Search
& \checkmark  & \checkmark &\checkmark & 2.01 & 52.7 \\
\rowcolor{gray!15}
 PrismRestore (Ours)
& \checkmark  & \checkmark & \checkmark
& \best{2.69} & \best{71.6} \\
\bottomrule
\end{tabular}%
% }
% \end{adjustbox}
\vspace{-2mm}
\end{table}
\noindent\textbf{Task-Specific Modulation Patterns.}
% To further examine the diversity of the searched priors,
% we analyze their modulation coefficients and channel-wise
% distributions in the latent stage. As shown in
% Fig.~\ref{fig:prior_distribution}, different restoration
% tasks exhibit distinct modulation patterns across channel
% intervals, both in the proportion of modulated channels
% and in the distribution of coefficient values.
% Notably, the defocus prior shows a different modulation
% profile from the motion-blur prior, while other tasks
% exhibit varying combinations of channel suppression
% and enhancement. These observations indicate that
% the offline search produces task-dependent modulation
% configurations rather than identical adjustments
% shared across degradation types.
To further examine the diversity of the searched priors,
we analyze their modulation coefficients and channel-wise
distributions in the latent stage. Across the five displayed channel intervals, the searched priors differ in both the proportion of modulated channels and the distribution of coefficient values (Fig.~\ref{fig:prior_distribution}). 
% As shown in Fig.~\ref{fig:prior_distribution}, different restoration tasks exhibit
% distinct modulation patterns across the selected channel
% intervals, both in the proportion of modulated channels
% and in the distribution of coefficient values.
% These observations indicate that the offline search
% produces task-dependent modulation configurations
% rather than identical adjustments shared across
% degradation types.

\begin{figure}[t]
    \centering
    \includegraphics[width=0.95\columnwidth]{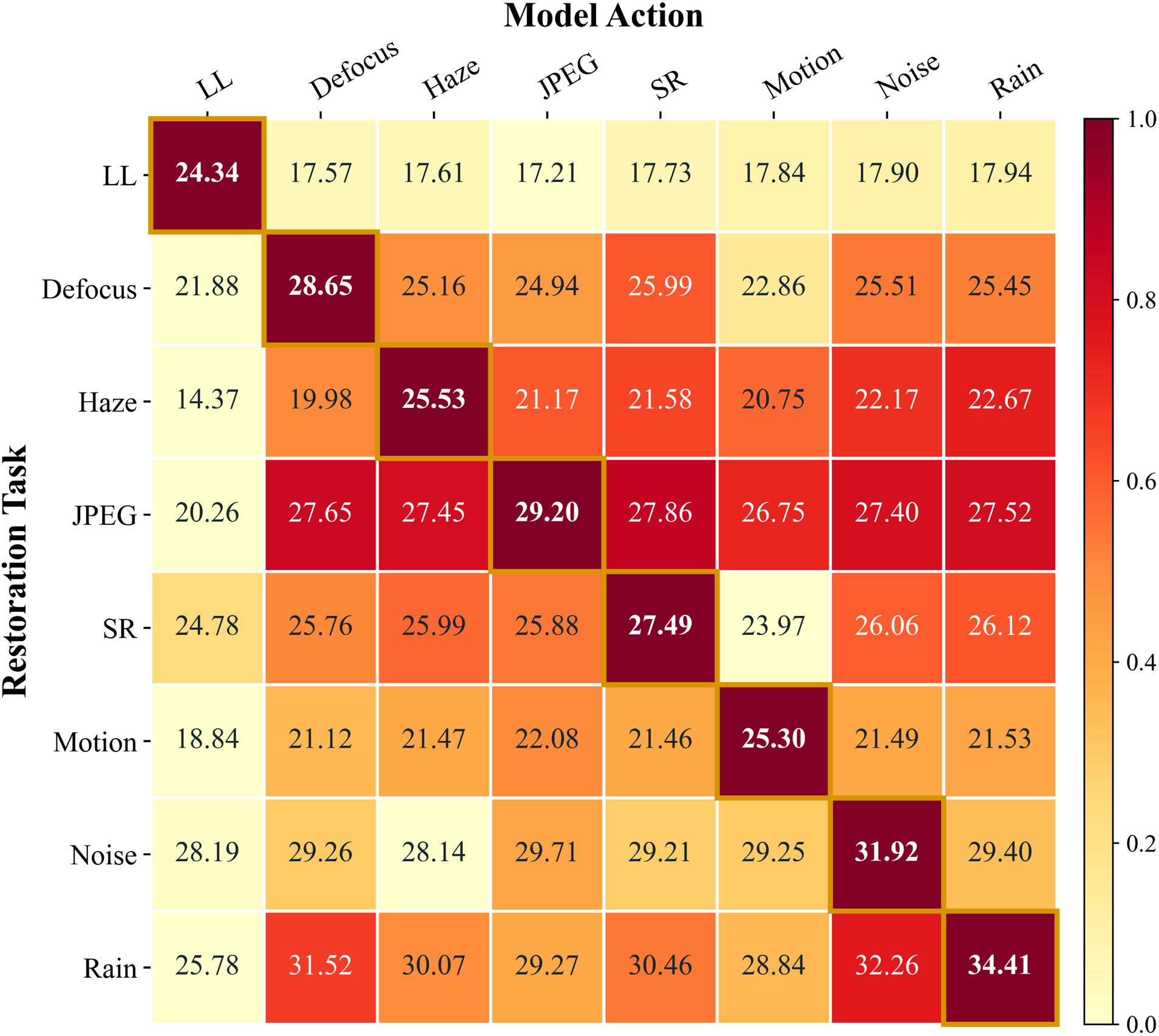}
    \caption{Task discriminability of the shared-backbone
restoration actions. Each cell reports the average
PSNR (dB), with highlighted diagonal entries
indicating task-matched actions. }
    \vspace{0mm}
    \label{fig:action_discriminability}
    
\end{figure}

\noindent\textbf{Task Discriminability of Restoration Actions.}
% To assess the task specificity of the constructed
% toolbox, we evaluate 8 restoration actions
% (Rain, Haze, Noise, LL, Motion, Defocus, JPEG, and SR)
% on 8 corresponding single-degradation test sets,
% each containing 100 images.
% As shown in Fig.~\ref{fig:action_discriminability},
% every task achieves its highest average PSNR with
% the corresponding action, yielding an average margin
% of 2.78 dB over the strongest mismatched action.
% Importantly, several off-diagonal responses remain
% competitive, indicating that the searched actions
% are not restricted to mutually exclusive restoration
% functions.
% Instead, task-conditioned modulation biases the
% shared backbone toward distinct degradation-specific
% behaviors while retaining cross-task restoration
% capabilities.
% Together with the component ablation, these results
% support the effectiveness of response-guided channel
% organization in constructing a compact yet
% distinguishable restoration toolbox.
We evaluate all eight actions across the
single-degradation test sets (Fig.~\ref{fig:action_discriminability}).
% We evaluate all eight actions on eight single-degradation test sets, each containing 100 images (Fig.~\ref{fig:action_discriminability}).
The task-matched action achieves the highest mean PSNR on every test set.
Averaged across tasks, its margin over the strongest mismatched action is 2.78~dB.
This matrix measures task preferences after averaging over images; the 71.6\% Match in Tab.~\ref{tab:toolbox_component} measures how often the matched action ranks first on individual images.

\noindent\textbf{Cross-Backbone Toolbox Construction.}
% We further evaluate the proposed toolbox
% construction method on three all-in-one
% restoration backbones: AdaIR~\cite{cui2025adair}, MoCE-IR~\cite{zamfir2025moceir}, and MiOIR~\cite{kong2024mioir}.
% As shown in Tab.~\ref{tab:cross_model},
% the constructed toolboxes achieve mean PSNR
% gains of 2.69, 2.83, and 2.47~dB, respectively,
% with Top-1 Match rates of 71.6\%, 69.4\%,
% and 63.6\%.
% These results indicate that task-specific
% restoration actions can be constructed across
% different all-in-one backbones, rather than
% being limited to AdaIR.
We further apply our task-prior search to three
all-in-one backbones: AdaIR~\cite{cui2025adair}, MoCE-IR~\cite{zamfir2025moceir}, and MiOIR~\cite{kong2024mioir}.
As shown in Tab.~\ref{tab:cross_model},
the resulting toolboxes achieve mean PSNR gains
of 2.69, 2.83, and 2.47~dB, with Top-1 Match
rates of 71.6\%, 69.4\%, and 63.6\%,
respectively. These results demonstrate
the applicability of our toolbox construction
method across different all-in-one backbones.

\begin{table}[!htb]
\centering
\footnotesize
% \vspace{-2mm}
\caption{Cross-backbone evaluation of the proposed task priors in terms of restoration gains and action consistency.}
% \begin{adjustbox}{width=\linewidth}
\label{tab:cross_model}
% \begin{adjustbox}{width=\linewidth}
% \setlength{\tabcolsep}{2.3pt}
\begin{tabular}{lcccc}
\toprule
\textbf{Method}&\textbf{Base(dB)} &\textbf{Base+P(dB)}& \textbf{Gain(dB)}& \textbf{Match(\%)}
% &\textbf{Runtime (s)}
\\
\midrule
% Greedy   
% &  & 
% & 21.42 & 4.804& \\
% Rollback   
% % &  &  
% & 21.42 & 4.804& \\
AdaIR  & 23.19 &25.88 &\best{2.69}&71.6\\
% & 0.14
% \rowcolor{gray!15}
MoCE-IR&   23.38 & 26.21& \best{2.83} &69.4 
% & 0.14
% \\

% Three& 2.76& 72.8& 28.42

\\
MiOIR & 22.76 &25.23 &\best{2.47}&63.6\\

\bottomrule
\end{tabular}
\vspace{-3mm}
\end{table}

%训练曲线
\subsection{Analysis of Policy Learning and Reward Design}
\label{sec:policy_reward_analysis}
% We conduct ablation studies on MiO100 Group A (640 images) to examine the contributions of policy initialization, GRPO optimization, reward design, and modulation refinement, as reported in Tab.~\ref{tab:policy_reward_ablation}.

% \noindent\textbf{Effects of policy initialization and optimization.}
% Removing SFT reduces PSNR from 21.19 to 19.83 dB, indicating that supervised initialization provides a useful starting point for sequential policy learning.
% Without GRPO, the SFT-only policy achieves 20.28 dB PSNR and 48.71 MUSIQ, compared with 21.19 dB and 54.24 for the full model. These improvements suggest that reinforcement learning plays an important role in refining action selection beyond supervised imitation, particularly in terms of perceptual quality.
% % Without GRPO, the SFT-only policy achieves
% % 20.28 dB PSNR and 52.84 MUSIQ, compared with
% % 21.19 dB and 54.24 for the full model.
% % These gains suggest that GRPO improves action
% % scheduling beyond the supervised trajectories.
% The final modulation refinement further increases
% PSNR from 21.03 to 21.19 dB and MUSIQ from
% 52.97 to 54.24, indicating that limited adjustment
% of the task priors can improve their compatibility
% with the learned policy.
% We examine the contributions of SFT initialization,
% GRPO optimization, and modulation refinement in
% Tab.~\ref{tab:policy_reward_ablation}.
% Removing SFT reduces PSNR from 21.19 to 19.83 dB,
% indicating that supervised initialization provides
% a useful starting policy for sequential restoration.

\noindent\textbf{Policy Initialization and Optimization.}
We evaluate the training variants on MiO100 Group A (Tab.~\ref{tab:policy_reward_ablation}).
Removing SFT initialization lowers PSNR from 21.19 to 19.83~dB.
The SFT-only variant reaches 20.28~dB PSNR and 48.71 MUSIQ; the full model improves these scores by 0.91~dB and 5.53 points, respectively.
Final refinement of the modulation priors adds 0.16~dB PSNR and 1.27 MUSIQ relative to the variant without refinement.

% Removing SFT initialization reduces PSNR from 21.19 to 19.83~dB, with consistent degradation in the
% remaining metrics. 
% This result indicates that supervised training provides an effective initialization for state-conditioned action selection before policy optimization.
% Conversely, removing RL yields the SFT-only policy, which obtains 20.28~dB PSNR and 52.84 MUSIQ. Subsequent RL optimization improves these results to 21.19~dB and 54.24, respectively, while also improving SSIM, LPIPS, MANIQA, and CLIP-IQA. 
% These results show that SFT establishes a useful initial policy, whereas RL further optimizes action selection according to restoration-quality feedback.
%质量效率散点图

\begin{figure}[t]
    \centering
    \includegraphics[width=\columnwidth]{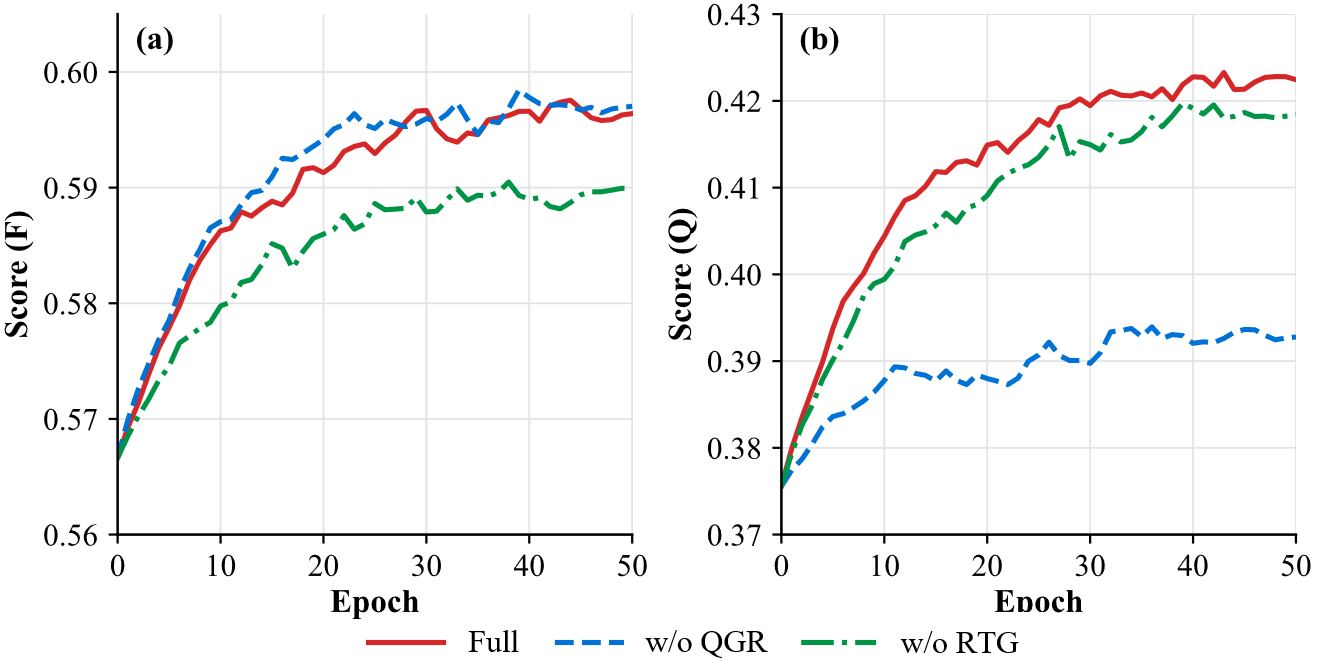}
    \vspace{-3mm}
    \caption{Training dynamics of quality feedback during GRPO optimization. (a) Reference-based quality score (F) and (b) no-reference quality score (Q) over 50 training epochs. }\vspace{-3mm}

    \label{fig:train_cve.png}
\end{figure}

\noindent\textbf{Effects of Quality Grouping and Process Feedback.} 
% Removing objective-wise normalization increases
% PSNR slightly from 21.19 to 21.27 dB and SSIM
% from 0.6826 to 0.6997, but reduces MUSIQ from
% 54.24 to 49.36 and worsens LPIPS from 0.3423
% to 0.3685.
% This contrast suggests that joint normalization
% shifts the optimization toward fidelity-oriented
% outcomes, whereas separate normalization better
% preserves perceptual quality with only a marginal
% PSNR trade-off.
% Removing process rewards also reduces PSNR to
% 20.81 dB and MUSIQ to 52.42, with declines
% across the remaining metrics.
% Unlike terminal-only feedback, the stepwise
% return-to-go incorporates intermediate quality
% changes into policy optimization, providing
% more differentiated supervision for actions
% along a restoration trajectory.
Removing quality-grouped normalization increases PSNR
slightly from 21.19 to 21.27 dB and SSIM from 0.6826
to 0.6997, but reduces MUSIQ from 54.24 to 49.36
and worsens LPIPS from 0.3221 to 0.3685.
This suggests that quality-grouped normalization better balances fidelity and perceptual quality, improving perceptual metrics at the cost of slightly lower PSNR and SSIM.
Removing process RTG decreases PSNR to 20.81 dB
and MUSIQ to 52.42, supporting the benefit of
incorporating intermediate quality changes into
policy optimization.
Fig.~\ref{fig:train_cve.png} further illustrates
these effects during GRPO training.
Without quality-grouped normalization, the reference-based
score $F$ remains comparable to the full model,
whereas the no-reference score $Q$ is generally lower throughout training.
Removing RTG leads to lower scores in both dimensions,
complementing the final restoration results in
Tab.~\ref{tab:policy_reward_ablation}.

% Without separately normalizing the fidelity and perceptual objectives, PSNR and SSIM slightly increase from 21.19~dB and 0.6826 to 21.27~dB and 0.6997, respectively. 
% However, perceptual quality deteriorates substantially: LPIPS increases from 0.3423 to 0.3685,
% while MANIQA, CLIP-IQA, and MUSIQ decrease from 0.3104, 0.4445, and 54.24 to 0.2761, 0.4113, and 49.36. 
% This behavior suggests that joint normalization favors fidelity-oriented optimization under the
% heterogeneous reward scales. 
% Separately normalizing the two objective groups therefore yields a substantially better perceptual result at a cost of only 0.08~dB in PSNR.

% Removing the process reward also degrades all six metrics, reducing PSNR from 21.19 to 20.81~dB and MUSIQ from 54.24 to 52.42.
% This consistent degradation indicates that terminal feedback alone
% provides relatively coarse credit assignment for intermediate actions.
% The return-to-go derived from stepwise quality changes supplies additional process-level supervision, enabling the policy to better distinguish beneficial from detrimental actions along a restoration trajectory.

\begin{table}[t]
\centering
\caption{
Ablation of policy learning and reward design on MiO100 Group A. The variants remove SFT initialization, GRPO optimization, quality-grouped normalization (QGR), process return-to-go (RTG), or final modulation refinement (Refine.), respectively.
}
\vspace{-2mm}
\footnotesize
\label{tab:policy_reward_ablation}
\setlength{\tabcolsep}{2.8pt}
\renewcommand{\arraystretch}{1.05}
% \small
\resizebox{\columnwidth}{!}{
\begin{tabular}{lccc
    @{\hspace{4pt}}c
    @{\hspace{4pt}}c
    @{\hspace{4pt}}c}
\toprule
\textbf{Setting}&
\textbf{~PSNR$\uparrow$} & \textbf{~SSIM$\uparrow$} & \textbf{~LPIPS$\downarrow$} &
\textbf{MANIQA$\uparrow$} & \textbf{CLIP-IQA$\uparrow$} & \textbf{MUSIQ$\uparrow$}\\
% Setting
% & PSNR
% & SSIM
% & LPIPS$\downarrow$
% & MANIQA
% & CLIP-IQA
% & MUSIQ \\
\midrule
w/o SFT
& 19.83 & 0.6476 & \second{0.3562} & \second{0.3068} & 0.4308 & 52.61 \\
w/o RL
& 20.28 & 0.6496 & 0.3943 & 0.2511 & 0.3907 & 48.71 \\
% \midrule
w/o QGR
& \best{21.27} & \best{0.6997}
& 0.3685 & 0.2761 & 0.4113 & 49.36 \\
w/o RTG
& 20.81 & 0.6654 & 0.3609 & 0.3016 & 0.4313 & 52.42 \\
w/o Refine
& 21.03 & 0.6794 & 0.3572 & 0.3052 &\second{0.4341}  &  \second{52.97}\\
% \midrule
\rowcolor{gray!15}
 PrismRestore (Ours)
& \second{21.19} & \second{0.6826}
& \best{0.3221}
& \best{0.3104}
& \best{0.4445}
& \best{54.24} \\
\bottomrule
\end{tabular}
}
\end{table}

\begin{figure}[t]
    \centering
    \includegraphics[width=\columnwidth]{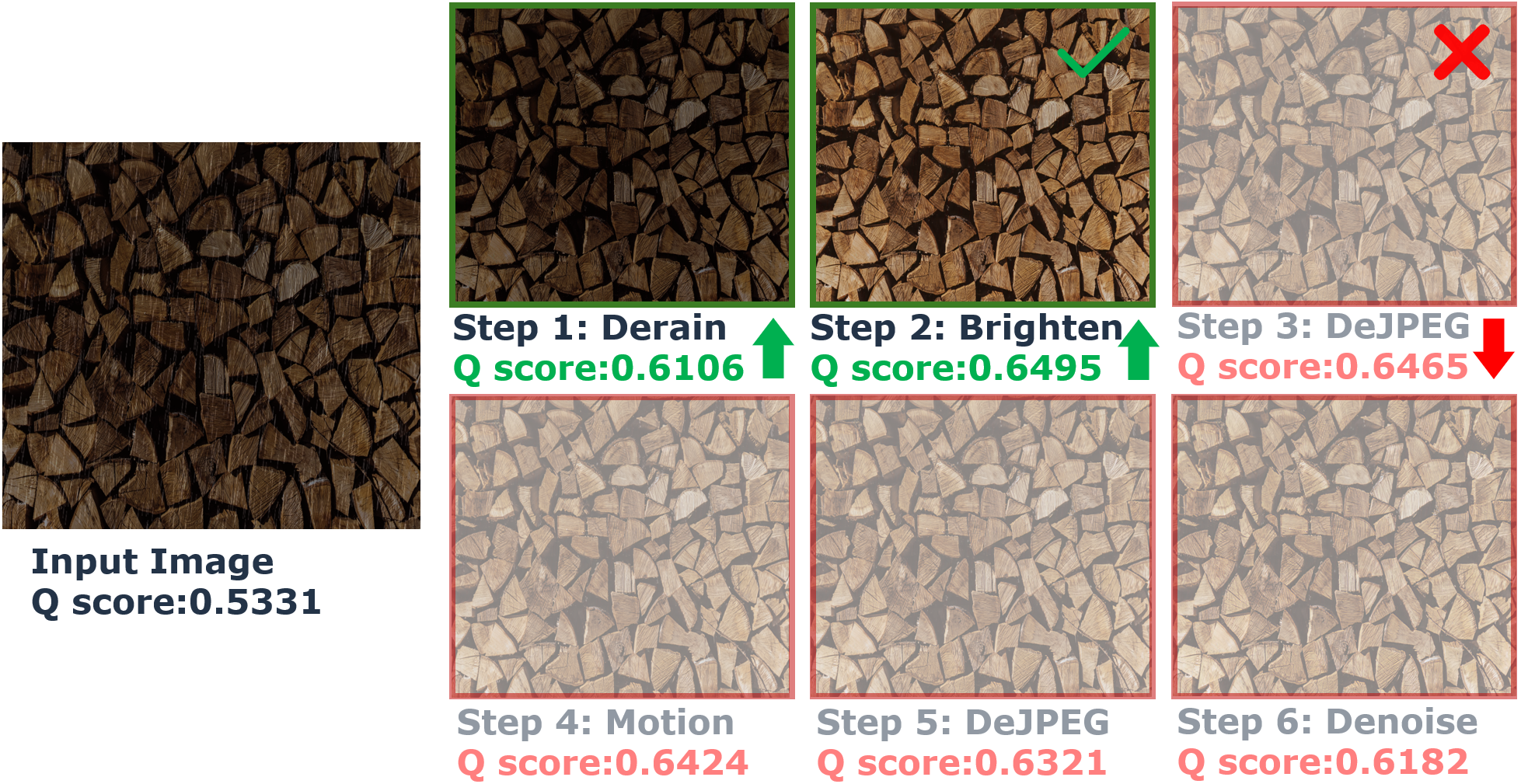}
    \vspace{-2mm}
    \caption{Sequential restoration with quality-guided stopping. The third update is rejected, while later steps show continued restoration without early stopping. }
\vspace{-3mm}
    \label{fig:tra_nan}
\end{figure}
\subsection{Analysis of Restoration Trajectories}
\noindent\textbf{Restoration Trajectory Analysis.} We visualize a representative restoration trajectory
in Fig.~\ref{fig:tra_nan} to examine how
the policy adapts action selection to different image
states. In this scene, deraining followed by
low-light enhancement increases the perceptual score
$Q$ from 0.5331 to 0.6495, while the third operation decreases \(Q\) and triggers early stopping. 
% In contrast, the
% second example benefits from a longer sequence of
% denoising and repeated deblurring, with $Q$ increasing
% from 0.2148 to 0.4349 before declining.
This trajectory illustrates how the learned policy selects complementary restoration actions, while quality-guided stopping prevents unnecessary subsequent operations once further restoration becomes detrimental.
% Together, these mechanisms enable effective sequential restoration without excessive tool invocations.
% % This observation indicates that restoration
% quality is not necessarily improved by additional
% actions and that the effective restoration depth need not be fixed. 
% The observed quality reversals
% also support the use of quality-guided stopping
% to avoid retaining detrimental restoration results.

\noindent\textbf{Execution Strategy Ablation.}
Tab.~\ref{execution} compares execution strategies on MiO100 Group A.
Applying the original AdaIR six times reduces PSNR from 19.07 to 17.03~dB, while random action selection achieves 18.61~dB.
Relative to greedy search, the learned policy improves PSNR by 0.37~dB and reduces LPIPS from 0.3635 to 0.3221.
It reduces the average number of tool calls from 29.65 to 3.26 and inference time from 38.96 to 4.67~s per image, corresponding to an 8.3$\times$ speedup.
% As shown in Tab.~\ref{execution}, repeatedly applying
% AdaIR degrades PSNR from 19.07 to 17.03~dB, suggesting
% that uncontrolled iterative restoration may cause
% over-restoration and error accumulation. Randomly
% selecting restoration actions also fails to improve PSNR over the original backbone. Greedy search
% improves PSNR to 20.82~dB but relies on immediate
% quality gains rather than optimizing long-term
% restoration outcomes, while incurring substantially
% more tool evaluations. In contrast, our learned
% policy achieves 21.19~dB PSNR with only 3.26
% restoration calls and 4.67~s per image,
% outperforming greedy search while being approximately
% 8.3 times faster.

\begin{table}[!htb]
\centering
\fontsize{8.5pt}{10pt}\selectfont
\caption{Ablation of execution strategies on MiO100 Group A.
Repeat.A applies the original AdaIR six times.
Calls denotes the average number of tool invocations
during inference.}
\vspace{-2mm}
\label{execution}
% \begin{adjustbox}{width=\linewidth}
% \setlength{\tabcolsep}{2.3pt}
\resizebox{\columnwidth}{!}{
\begin{tabular}{lccccc}
\toprule
\textbf{Strategy}& \textbf{PSNR$\uparrow$}& \textbf{SSIM$\uparrow$}& \textbf{LPIPS$\downarrow$}&\textbf{Calls}&\textbf{Time(s)}
% &\textbf{Runtime (s)}
\\
\midrule
% Greedy   
% &  & 
% & 21.42 & 4.804& \\
% Rollback   
% % &  &  
% & 21.42 & 4.804& \\
AdaIR &19.07 &0.6063&0.4162 &-- & 1.36
% & 0.14
\\
Repeat.A&17.03 &0.6198&0.4159 &6.00 & 5.84
% & 0.14
\\
Random& 18.61&0.6287&0.4060 & 3.46&3.05 
% & 0.14
\\
Greedy& \second{20.82}&\second{0.6602}&\second{0.3635}&29.65&38.96
% & 0.15
\\
\rowcolor{gray!15}
 PrismRestore (Ours)& \best{21.19}&\best{0.6826}& \best{0.3221}& 3.26&4.67

\\

\bottomrule
\end{tabular}}
\vspace{-4mm}
% \end{adjustbox}
\end{table}

%% file: sec/5_conclusion.tex
\section{Conclusion}
We presented PrismRestore, a compact agentic image
restoration framework built upon a shared-backbone
toolbox. 
By leveraging task-specific channel
modulation, our offline prior search constructs
multiple callable restoration actions from a
single frozen all-in-one model, substantially
reducing toolbox storage overhead. 
A GRPO-trained
policy enables adaptive action composition
according to evolving image states, with
quality-grouped and process-aware rewards
supporting sequential optimization.
Extensive
experiments on synthetic and real-world benchmarks
demonstrate improved restoration performance
over the evaluated all-in-one baselines and
competitive restoration quality relative to
multi-model agents, alongside reduced toolbox
storage and improved inference efficiency.

\noindent\textbf{Limitations.}
While PrismRestore achieves competitive
restoration quality with minimal task-specific
parameters, further improving perceptual
quality without increasing toolbox storage
overhead remains an important direction.
Moreover, the toolbox is built for predefined
degradations, and extending it to unseen ones
may require additional prior construction
and policy adaptation.
% \noindent\textbf{Limitations.}
% Despite strong reconstruction fidelity, PrismRestore trails
% some agentic baselines on perceptual quality metrics.
% Moreover, the no-reference score used for stopping may not
% always align with human preferences.
% Future work could explore perceptually oriented adaptation
% and policy objectives while retaining a compact toolbox,
% along with uncertainty-aware quality assessment and
% preference-based feedback for more reliable stopping.

%% file: sec/LLM_Usage_Statement.tex
\section*{LLM Usage Statement}
We used ChatGPT (OpenAI) to polish selected passages, including related
work, adjust the layout of some figures and tables, and edit parts of
the implementation code. We also used ChatGPT to obtain reviewer-style
feedback on the full manuscript and supplementary material.